\documentclass{article}
\usepackage{ijcai20}
\usepackage{footnote}
\usepackage{threeparttable}
\usepackage{times}
\usepackage{soul}
\usepackage{url}
\usepackage[hidelinks]{hyperref}
\usepackage[utf8]{inputenc}
\usepackage[small]{caption}
\usepackage{graphicx}
\usepackage{amsmath}
\usepackage{amsthm}
\usepackage{booktabs}
\usepackage{algorithm}
\usepackage{algorithmic}

\usepackage{times}
\usepackage{epsfig}
\usepackage{amssymb}
\usepackage{amsfonts}
\usepackage{mathtools}
\usepackage{mathabx}
\usepackage{bm}
\usepackage{mathrsfs}
\usepackage{subfigure}
\usepackage{multirow}
\usepackage{booktabs}
\usepackage{color}
\usepackage{epstopdf}
\usepackage{graphicx}
\usepackage{float}

\title{Recurrent Relational Memory Network for Unsupervised Image Captioning}
\author{
Dan Guo
\and Yang Wang$^{*}$
\and Peipei Song\thanks{Corresponding authors.}
\And Meng Wang
\affiliations
Key Laboratory of Knowledge Engineering with Big Data (HFUT), Ministry of Education\\
School of Computer Science and Information Engineering, Hefei University of Technology (HFUT)\\
\emails
\{guodan, yangwang\}@hfut.edu.cn, \{beta.songpp, eric.mengwang\}@gmail.com
}

\begin{document}

\maketitle

\begin{abstract}
Unsupervised image captioning with no annotations is an emerging challenge in computer vision, where the existing arts usually adopt GAN (Generative Adversarial Networks) models. In this paper, we propose a novel memory-based network rather than GAN, named Recurrent Relational Memory Network ($R^2M$). 
Unlike complicated and sensitive adversarial learning that non-ideally performs for long sentence generation, $R^2M$ implements a concepts-to-sentence memory translator through two-stage memory mechanisms: fusion and recurrent memories, correlating the relational reasoning between common visual concepts and the generated words for long periods.
$R^2M$ encodes visual context through unsupervised training on images, while enabling the memory to learn from irrelevant textual corpus via supervised fashion. Our solution enjoys less learnable parameters and higher computational efficiency than GAN-based methods, which heavily bear parameter sensitivity. We experimentally validate the superiority of $R^2M$ than state-of-the-arts on all benchmark datasets.
\end{abstract}

\section{Introduction}


\begin{figure}
    \centering
    \includegraphics[width=\linewidth]{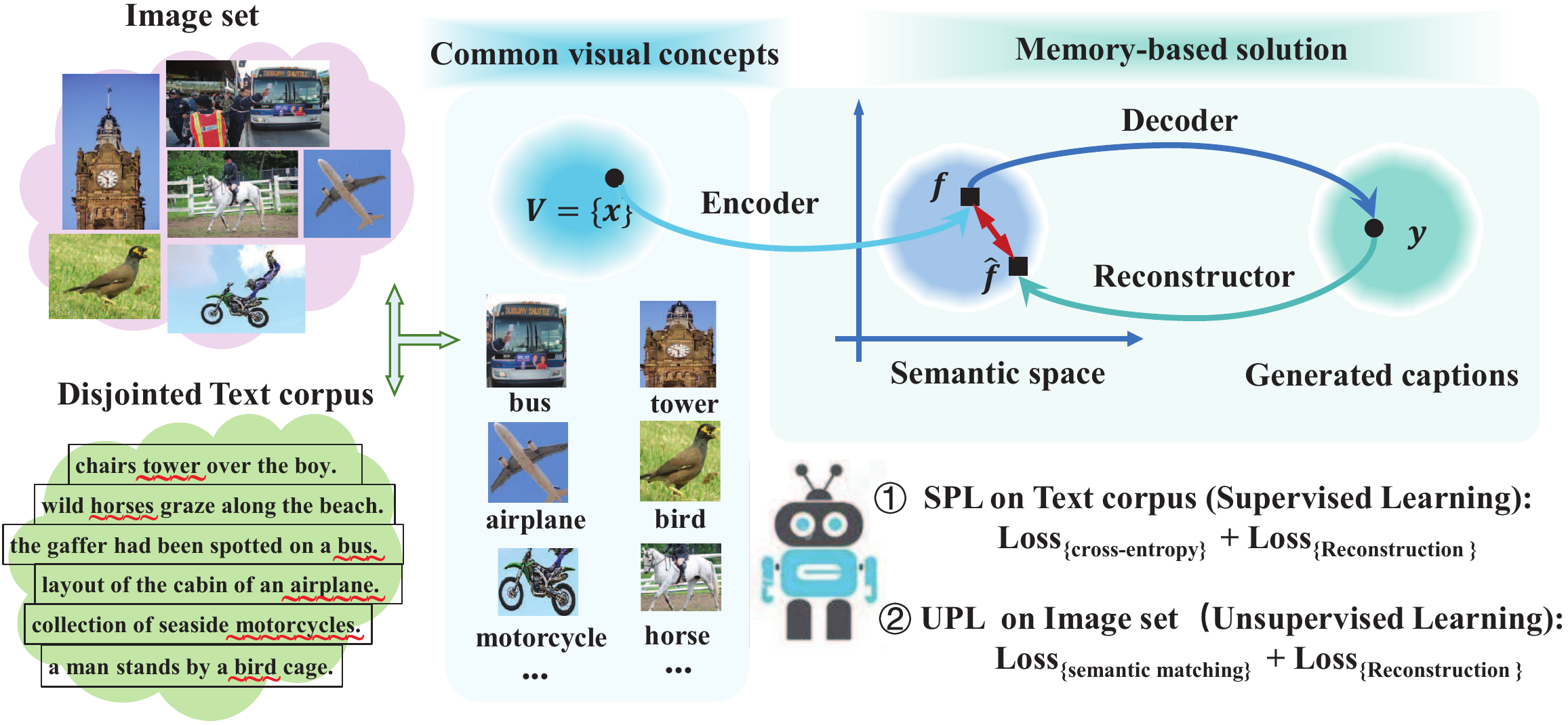}\\
    \caption{The basic idea of our solution; that is to learn common concepts co-occurred in both image set and text corpus. We propose a light memory network named $R^2M$ to memorize and translate concepts to a sentence. 
    $R^2M$ first imitates humans to listen to serval salient words and make sentences, exhibiting supervised learning (SPL) on text corpus. Then it locates visual concepts in images and makes sentences, representing unsupervised learning (UPL). The proposed memory demonstrates effective semantic reasoning for sequential learning.
    }
  \label{fig_2}
\end{figure}

Traditional image captioning \cite{yao2019hierarchy,huang2019attention} requires full supervision of image-caption pairs annotated by humans. However, such full supervision is ridiculously expensive to acquire in cross-modal datasets. Recently, substantial researches tend to flexible constrained caption tasks, such as unpaired captioning \cite{gu2019unpaired,guo2019mscap} and unsupervised captioning \cite{feng2019unsupervised} with weak or no supervised cues. It is challenging to leverage the independent image set and sentence corpus to train a reliable image captioning model; worse still, image captions usually cover specified or insufficient topics, \emph{e.g.,} the well-known benchmark MSCOCO images~\cite{lin2014microsoft} just cover 80 object categories, raising up the challenges to generate rich semantical and grammatical sentences.
\begin{figure*}
   \centering
   \includegraphics[width=\linewidth]{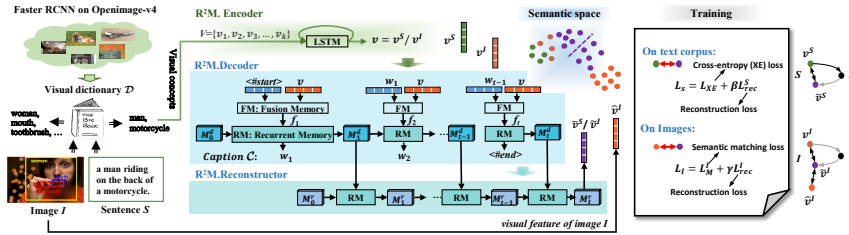}
   \caption{An overview of $R^2M$ (\underline{R}ecurrent \underline{R}elational \underline{M}emory network).
   We perform unsupervised captioning through mess occurrences of common visual concepts in disjoint images and sentences. A visual ${\mathcal{D}}ictionary$ ${\mathcal{D}}$ built upon Openimage-v4 is utilized to filter out crucial visual concepts in image $I$ or sentence $S$. 
   Inheriting from the encoded vector of visual concepts ${\bm v}={\bm v}^I$ or ${\bm v}^S$, $R^2M. Decoder$ generates ${\mathcal{C}}aption\ \mathcal{C}=\{w_1,\cdots, w_t\}$ by leveraging a two-stage memory mechanism, while $R^2M. Reconstructor$ recalls the memory for rebuilding visual semantics $\tilde { v}^I$ or $\tilde { v}^S$ corresponding to respective ${\mathcal{C}}$. The supervised loss $L_{S}$ on text corpus is optimized via cross-entropy loss $L\!_{X\!E}(S,C)$ and reconstruction loss $L_{rec}^{S}({\bm v}^S, \tilde {\bm v}^S)$. Then unsupervised loss $L_{I}$ for images is optimized by semantic matching (triplet semantic ranking) loss $L^I_M(\hat {v}^I, \tilde { v}^I)$ and reconstruction loss $L_{rec}^{I}({ v}^I, \tilde { v}^I)$.
  }
\label{fig_3}
\end{figure*}

There are merely two unsupervised methods test on the disjointed image and text corpus data.
\cite{feng2019unsupervised} proposed an architecture comprising of an image encoder, a sentence generator, a discriminator with adversarial loss and concept reward. \cite{laina2019towards} learned a joint semantic embedding space, where either images or sentences were transformed. A discriminator was then designed to judge
where the embedding feature came from, image or sentence
domain. Both of them resolved the task with adversarial training, while obeying the usage of GAN (Generative Adversarial Networks) in unsupervised mode \cite{lample2017unsupervised,donahue2019large,yang2018unsupervised}.
As it is widely known, current GAN methods
based on ordinary recurrent models (\emph{e.g.}, LSTM)
always employ RL heuristics and are quite sensitive to parameter initializations and hyper-parameter
settings \cite{nie2018relgan}. 

Orthogonal to above GAN-based models, in this paper, we propose a novel memory-based solution, named \underline{R}ecurrent \underline{R}elational \underline{M}emory Network ($R^2M$). The novelty of $R^2M$ lies in its exploitation on visual concepts and describing image via memory, serving as a concepts-to-sentence memory translator to learn the textual knowledge from discrete common concepts in diverse sentences, meanwhile being capable of making sentences with correctly semantic and grammar syntax rules.

As illustrated in Fig.\ref{fig_2}, $R^2M$ explores the latent relevant semantic learning with the memory network, so as to enjoy the flexible and augmented memory capacity for both vision and natural language processing tasks \cite{pei2019memory,fan2019heterogeneous}.
Our intuition is that memory is proficient at storing and retrieving relational contexts to correlate input information \cite{huang2019acmm}, while inhibits vanishing gradients \cite{santoro2018relational,fu2019non}. 
As illustrated in Fig.~\ref{fig_3}, $R^2M$ consists of two-stage memories, \emph{fusion memory} (FM) and \emph{recurrent memory} (RM). The relational reasoning based on FM and RM in our captioning process, not only considers the dependencies between words and common concepts, but also distills the useful context into the memory, retaining for long periods. Technically, $R^2M$ performs the recurrent relational reasoning through multi-head self-attention and a relational gate (detailed in Section~\ref{subsec:2.2}).


So far, FM and RM memories resolve the relational reasoning for text generation. 
As inspired, as shown in Figs.\ref{fig_2} and \ref{fig_3}, we develop a joint exploitation of supervised learning (SPL) and unsupervised learning (UPL) on the disjoint datasets. In particular, the SPL strategy is to learn the memories and make sentences from several salient words that separately appeared in text corpus, while the UPL training is to fine-tune the memories and make much more appropriate sentences about the visual context in the image. The supervised training on text corpus incorporates inductive semantic bias into the language model training. Turning to visual concepts in images without supervision cues, we explore a semantic matching (hinge-based triplet ranking) loss $L^I_M$ and reconstruction loss. These two losses encourage the cross-modal similarity score of image,  along with the generated sentence to be larger than that of the other sentences. For $L^I_M$, we distinguish the negatives $({ I'}, \bm{\mathcal{C}}_I)$ and $(I, {\bm{\mathcal{C'}}})$ from a positive image pair $(I, \bm{\mathcal{C}}_I)$, which is discussed in Section~\ref{subsec:training}.

The major contributions are summarized as follows:
  \begin{itemize}
  \item Orthogonal to GAN-based architectures 
        for unsupervised image captioning, we propose a novel light Recurrent Relational Memory Network ($R^2M$), which merely utilizes the attention-based memory (detailed in Section~\ref{subsec:2.2}) 
        to perform the relational semantics reasoning and reconstruction.
  \item A joint exploitation of Supervised learning on text corpus and Unsupervised learning on images is proposed. We optimize the cross-modal semantic alignment and reconstruction via an unsupervised manner to achieve a novel concepts-to-sentence translation.
  \item The proposed $R^2M$ achieves better performances than state-of-the-arts on all the current unsupervised datasets: MSCOCO paired Shutterstock captions, Flickr30k paired MSCOCO captions and MSCOCO paired GCC (Google’s Conceptual Captions).
  \end{itemize}

\section{Proposed Method}
\label{meth}
In this section, we formally discuss our proposed $R^2M$.
The overall architecture of  $R^2M$ is depicted in Fig.\ref{fig_3}, which consists of three modules: \emph{encoder}, \emph{decoder} and \emph{reconstructor}.

\subsection{R$^2$M. Encoder}
We first discuss the encoder. 
A visual dictionary $\bm{\mathcal{D}}$ is learned ahead by using Faster R-CNN \cite{huang2017speed} trained on a public dataset OpenImages-v4 \cite{krasin2017openimages,kuznetsova2018open} to cover the majority of common visual concepts in daily conversations, which is used to filter out visual concepts $V=\{v_{i}|_{i=1}^k\}$ of image $I$ or sentence $S$. 
After that, visual concepts $V$ are randomly and sequentially incorporated into LSTM with their word embeddings, leading to the encoded vector ${\bm v}={\bm v}^I$ or ${\bm v}^S$ from $I$ or $S$.


\subsection{R$^2$M. Decoder}
\label{subsec:2.2}
Details of decoder are illustrated in Fig.\ref{fig_4}.
The effect of $R^2M. Decoder$ is to generate grammatical and semantical sentences from a few discrete words, \emph{e.g.}, translating ``\emph{man}” and ``\emph{motorcycle}” to ``\emph{a man riding on the back of a motorcycle}”. The set of visual concepts has no available grammar and syntax contexts. Based on that, we train the model to \textit{think, infer} and \textit{talk} about as human beings. To address this issue, we propose a memory-based decoder, 
which not only considers the correlation between visual concepts and current generated word, but also captures the temporal dependencies and distills the underlying memory information.

\subsubsection{Relation Learning I: Fusion Memory (FM) }
\begin{figure}[t]
   \centering
   \includegraphics[width=3.2in]{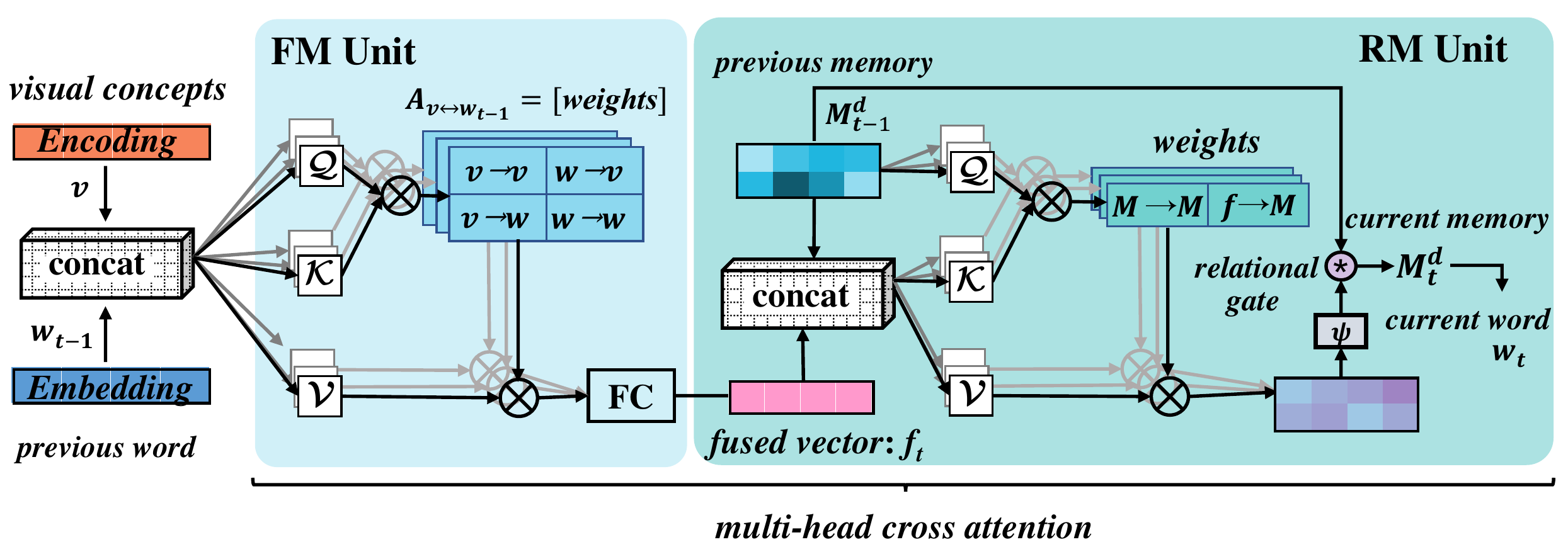}
   \caption{Memory mechanism in $R^2M. Decoder$. For the $t$-th step, FM fuses the encoded vector of visual concepts $v$ and the previous word $w_{t-1}$ to a semantic vector $f_t$, and the RM unit updates memory $M^{d}_{t}$ by incorporating [$M^{d}_{t-1}$,$f_t$]. Both of them indicate the multi-head cross attention among different semantic variables.}
\label{fig_4}
\end{figure}

As shown in Fig.\ref{fig_4}, the fusion memory (FM) in the decoder phase is used to learn the relationship between visual concepts and generated words, while recurrent memory (RM) in both decoder and reconstructor recurrently updates the memory to deliver useful semantics.
At time step $t$, FM learns the implicit relationship between the encoded concept vector ${\bm v} \in \mathbb{R}^{d}$ and previous generated word $\bm {w}_{t-1} \in \mathbb{R}^d$. We adopt a row-wise concatenation to acquire a joint feature matrix ${\bf x_{t}} = [{\bm v};{\bm {w}_{t-1}}] \in \mathbb{R}^{2\times d}$, upon which multi-head self-attention \cite{vaswani2017attention} is performed. 
The intuition is to explore the correlation between ${\bm v}$ and $\bm {w}_{t-1}$. 
We consider the influences: ${\bm v} \rightarrow \bm {v}$, ${\bm v} \rightarrow \bm {w}_{t-1}$, ${\bm {w}_{t-1}} \rightarrow \bm {w}_{t-1}$ and ${\bm w}_{t-1} \rightarrow \bm {v}$. They are performed by the dot-product of query and key transformers of ${\bf x_{t}}$ as follows:
\begin{equation}
\begin{split}
   A_{v \leftrightarrow w_{t-1}}&=\begin{bmatrix}
   \bm v\rightarrow \bm v, & \bm w_{t-1}\rightarrow \bm v\\
    \bm v\rightarrow \bm w_{t-1}, & \bm w_{t-1}\rightarrow \bm w_{t-1}
    \end{bmatrix}\in \mathbb{R}^{2\times 2}\\
   &=\text{softmax}\big(\underbrace{{\bf x_{t}}U_q}_{\text{query}}\cdot (\underbrace{{\bf x_{t}}U_k}_{\text{key}})^{\top }/\sqrt{\lambda_1}\big),
\end{split}
\end{equation}
where $U_q
, U_k \in \mathbb{R}^{d\times d_k}$ are parameters of linear transformations of ${\bf x_{t}}$ ($query$ and $key$); $\lambda_1$  is a scaling factor to balance the fusion attention distribution.

The cross interaction between ${\bm v}$ and $\bm {w}_{t-1}$ is calculated based on both attended weights and values 
as follows:
   \begin{equation}
   \tilde{\bf x}_{t}=A_{v \leftrightarrow w_{t-1}}\cdot \underbrace{ ([{\bm v};{\bm w_{t-1}}] \cdot U_{v})}_{\text{value}} \in \mathbb{R}^{2\times d_v},
   \end{equation}
where $U_v\in \mathbb{R}^{d\times d_v}$ is another learnable parameter of linear transformations of ${\bf x_{t}}$ ($value$).

To ensure diverse and comprehensive attention guidance, we fuse ${\bm v}$ and ${\bm w_{t-1}}$ from $H$ perspectives. There are $H$ heads of independent attention executions. The outputs are concatenated into a new matrix ${\bf x'}_{t}$ as follows:
   \begin{equation}
   {\bf x'}_{t}=\big[{\tilde{\bf x}^h_{t}}\big]\big|\big|_{h=1}^{H} =[{\tilde{\bf x}^1_{t}}, \cdots, {\tilde{\bf x}^H_{t}}] \in \mathbb{R}^{2\times (H\cdot d_v)},
   \end{equation}
where $||$ denotes column-wise concatenation. Finally, we use a fully-connection (linear) layer to convert the matrix ${\bf x'}_{t}$ into a fusion-aware feature ${\bm f_{t}}$ below:
   \begin{equation}
   {\bm f_{t}}=FC({\bf x'}_{t}) \in \mathbb{R}^d.
   \end{equation}

\subsubsection{Relation Learning II: Recurrent Memory (RM)}
\begin{figure}[t]
   \centering
   \includegraphics[width=2.8in]{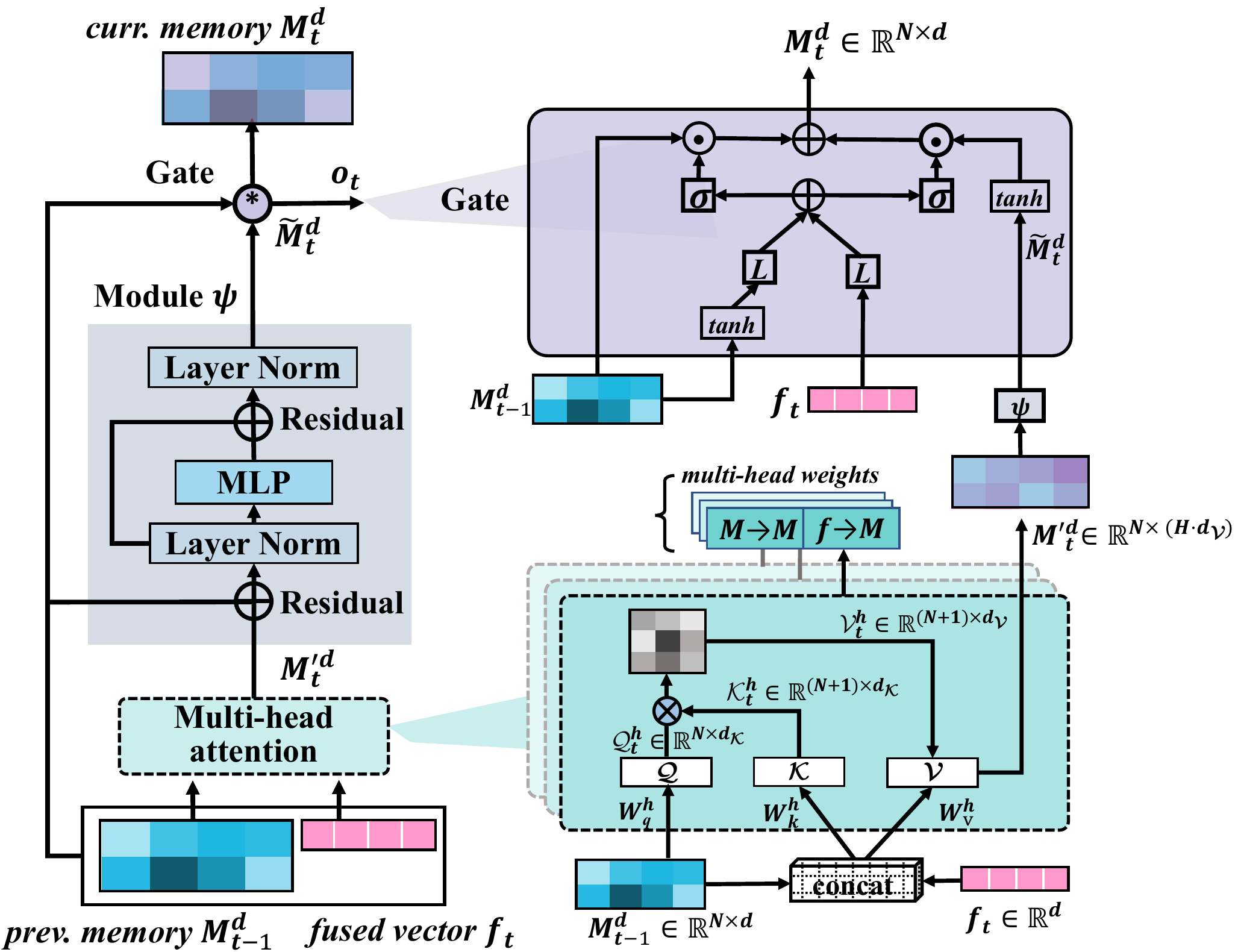}
   \caption{Relational gate in the RM unit of  $R^2M. Decoder$. Turning to $R^2M. Reconstructor$, based on the same RM unit, we incorporate the previous reconstructed memory $M_{t-1}^r$ and current decoded vector $M^{d}_{t}$ to learn current reconstructed memory $M_{t}^r$. Here, $\odot$, $\sigma$, and $L$ denote dot product, sigmoid activation and a fully-connection (linear) layer, respectively. 
   }
\label{fig_5}
\end{figure}

Observing ${\bm {f}_{t}}$  at the $t$-th time step, RM recurrently learns a decoded memory variable ${M}_{t}^{d}$ as shown in Fig.\ref{fig_5}. To distill the information worthy to retain in memory, we apply a relational gate for the recurrent memory updating among sequential learning. First, the multi-head self-attention is recycled to model latent transformers of previous memory state $M^{d}_{t-1}$ and fusion-aware feature ${\bm {f}_{t}}$
, where $M^{d}_0$ is initialized with zero-padding.
Note that we merely focus on the memory variation itself. The $query$ is related to $M^{d}_{t-1}$, $key$ and $value$ refer to $[M^{d}_{t-1};{\bm f_{t}}]$, implying that the joint effect of $[M^{d}_{t-1};{\bm f_{t}}]$ is learned under the guidance of $M^{d}_{t-1}$. In this part, the detailed dimensions of parameters are shown in Fig.\ref{fig_5}.
\begin{equation}
\label{rm}
\begin{split}
   A^h_{\bm f_{t}\rightarrow {\bm M^{d}_{t-1} }}&=\![
   \bm M^{d}_{t-1}\!  \rightarrow\!  \bm M^{d}_{t-1}, {\bm f\! _{t}} \!  \rightarrow \! \bm M^{d}_{t-1}
   ]^h\\
   &=\! \text{softmax}\! \big( \underbrace{{\! {\bm M\! _{t-1}^{d}}\!  } W\! _q^h}_{\text{query}}\! \cdot (\underbrace{{[\! {\bm M\! _{t-1}^{d}};{\bm f\! _{t}}\! ]}W\! _k^h}_{\text{key}})\! ^{\top }\! /\! \sqrt{\lambda_2}\big),
\end{split}
\end{equation}
\begin{equation}
{M'}_{t}^{d}=\Big[ A^{h}_{\!{\bm f\! _{t}} \rightarrow\! M^{d}_{t-1}  }\! \cdot\!  \Big (\! \underbrace{[{\bm M\! _{t-1}^{d}};{\bm f\! _{t}}]W\! _{v}^{h}}_{\text{value}}\Big )\! \Big]\big|\big|_{h=1}^{H}
\end{equation}
where $W^h_q
,W^h_k\in \mathbb{R}^{d\times d_{\mathcal{K}}}$ and $W^h_v\in \mathbb{R}^{d\times d_{\mathcal{V}}}$ are learnable parameters, and $\lambda_2$ is the scaling factor to balance the attention distribution in RM.

\textbf{Module} ${\bm \psi}$ ${M'}_{t}^{d}$ is then fed into two residual connection layers and one row-wise multi-layer perception (MLP) with layer normalization. Thus, we achieve a memory gain $\tilde {M}_{t}^{d}$.

\textbf{Relational Gate} To model the temporal dependencies along the adjacent memories, we update the memory state in a gated recurrent manner. Specifically, we apply input gate $g_{i,t}$ and forget gate $g_{f,t}$ to balance the memory updating from the current memory gain $\tilde{M}_{t}^{d}$ and original memory $M^{d}_{t-1}$, respectively. Both $g_{i,t}$ and $g_{f,t}$ are affected by  ${\bm {f}_{t}}$ and $M^{d}_{t-1}$.
\begin{equation}
   \label{gating}
   \left\{\begin{split}
   g_{i,t} &= \sigma (W_i\cdot{\bm f_{t}}+U_i\cdot \text{tanh}(M^{d}_{t-1})+b_i)\\
   g_{f,t} &= \sigma (W_f\cdot{\bm f_{t}}+U_f\cdot \text{tanh}(M^{d}_{t-1})+b_f)\\
    M^{d}_{t}&= g_{i,t}\odot \text{tanh}(\tilde{M}_{t}^{d})+ g_{f,t}\odot M^{d}_{t-1},
   \end{split}\right.
\end{equation}
where $\odot$ and $\sigma$ denote dot product and sigmoid functions.
Based on the updated memory $M^{d}_{t}$, RM outputs the word $w_{t}$:
   \begin{equation}
    w_{t} = argmax \{ \text{softmax}(W_{d} \cdot M^{d}_{t}) \},\\
   \end{equation} 
where $W_{d}$ is a learnable matrix that maps $M_{t}^{d}$ to a vector with the dictionary size.


\subsection{R$^2$M. Reconstructor}
So far, the decoder yields a pipeline to translate discrete visual concepts into a formal sentence. Here, we attempt to ensure that $R^2M$ can talk about correct contents. As inspired, we reversely reconstruct the concept semantics, \emph{i.e.}, 
rebuilding the crucial concept semantics from the generated sentence. 
We adopt the memory unit RM to compose the $R^2M. Reconstructor$. Note that learnable parameters of RM in $R^2M. Decoder$ and $Reconstructor$ are completely different.

If we define the RM operation in $R^2M. Decoder$ as a function $M^{d}_{t}= RM (M^{d}_{t-1}, f_t)$ involving Eqs. \ref{rm}$\sim$\ref{gating}, the $R^2M. Reconstructor$ operation is formulated as follows:
\begin{equation}
   M^{r}_{t}=RM(M^{r}_{t-1}, M_{t}^{d}), t\in\{0,\cdots,len\},
\end{equation}
where $M^{r}_t$ indicates a reconstructed memory at time $t$, $M^{r}_0$ is initialized with zero-padding, and $len$ is the length of the generated caption $\bm{\mathcal{C}}$. The last output of $R^2M. Reconstructor$ is treated as the reconstructed vector of concepts, denoted as $\tilde {\bm v}^{I}$ or $\tilde {\bm v}^{S}$ corresponding to image $I$ or sentence $S$.

\subsection{Training}
\label{subsec:training}
\subsubsection{Supervision Learning on Text Corpus}
We train the concepts-to-sentence decoder $R^2M. Decoder$ by maximizing log-likelihood of the generated sentences with original corpus sentences:
\begin{align}
   L\!_{X\!E}=-\sum_{t=1}^{len}\text{log} p(w_{t}|w_{t-1}).
\end{align}

For $R^2M. Reconstructor$, there is the reconstructed vector $\tilde {\bm v}^{S}$ corresponding to 
sentence $S$. We align it in an unsupervised mode. The full objective on text corpus is:
\begin{equation}
   L_{S}= L\!_{X\!E} + \beta L_{rec}^{S},
\end{equation}
where $L_{rec}^{S}= ||{\bm v}^{S}-\tilde {\bm v}^{S}||_{L2}^{2}$, $\beta $ is a hyper-parameter, and $||.||_{L2}^{2}$ denotes the L2-norm loss.

\subsubsection{Unsupervised Visual Alignment on Images}
The remaining question is how to achieve a better generalization ability with no supervision cues for image captioning? To answer this question, we adopt a hinge-based triplet ranking loss $L^I_M$, which encourages the semantic relevance of $(I, \bm{\mathcal{C}}_I)$ to be much larger than other negative examples. We choose the hardest negatives ${I'}$ and ${ \bm{\mathcal{C'}}}$ for a positive pair $(I, \bm{\mathcal{C}}_I)$, and perform  $L^I_M$ as follows:
\begin{equation}
   \begin{split}
   L\!_{M}^{I}=&[\mathbf m-\mathcal S(I, \bm{\mathcal{C}}_I)+\mathcal S({ I'}, \bm{\mathcal{C}}_I)]_{+}+\\
   &[\mathbf m-\mathcal S(I, \bm{\mathcal{C}}_I)+\mathcal S(I, { \bm{\mathcal{C}}'})]_{+} \\
   s.t., \ &{I'}={\underset{I'\neq I}{\text{argmax}}}\mathcal S( I', \bm{\mathcal{C}}_I), {\bm{\mathcal{C}}'}=\underset{I'\neq I}{\text{argmax}}\mathcal S( I, \bm{\mathcal{C}}_{I'}),
   \end{split}
   \label{eq:12}
\end{equation}
where $[x]_{+} = \text{max}(x,0)$, $\mathcal S(\cdot)$ is the similarity function calculated by inner product, and $\mathbf m$ serves as a margin parameter. $\mathcal S(I, \bm{\mathcal{C}}_I)= \mathcal S(\hat {\bm v}^I, \tilde {\bm v}^{I})$, where $\hat {\bm v}^I$ is the visual feature of image $I$ extracted by Inception-V4 \cite{szegedy2017inception} and $\tilde {\bm v}^{I}$ is the reconstructed vector by the RM unit. For computational efficiency, we search the negatives $I'$ and $\bm{\mathcal{C'}}$ within each mini-batch instead of 
the entire training set.

Besides, the image reconstruction loss $L_{rec}^{I}$ is utilized to train the model. The full objective on images is:
\begin{equation}
   L_{I} = L^I_M + \gamma L_{rec}^{I},
  \end{equation}
where $L_{rec}^{I}= ||{\bm v}^{I}-\tilde {\bm v}^{I}||_{L2}^{2}$ and $\gamma $ is a hyper-parameter.

\begin{table*}[t]
   \setlength{\tabcolsep}{.3em}
   \centering
   \scalebox{0.9}{
   \begin{tabular}{c|c|cccccccc}
      \toprule[1pt]
            Dataset &\multirow{2}*{Method }  &   \multirow{2}*{B-1}   &   \multirow{2}*{B-2 }  &  \multirow{2}*{ B-3 }  &  \multirow{2}*{ B-4 }  & \multirow{2}*{METEOR} &  \multirow{2}*{ ROUGE }   &  \multirow{2}*{ CIDEr }   &   \multirow{2}*{SPICE } \\ \cline{1-1}
           Images$\leftrightarrow$Captions  &~ \\ \midrule
      \multirow{2}*{MSCOCO$\leftrightarrow$Shutterstock} &UC-GAN~\cite{feng2019unsupervised} &     41.0     &    22.5    &    11.2    &    5.6     &  12.4  &    28.7    &    28.6    &    8.1    \\
                  ~   &$R^2M$                  &    {\bf 44.0 }   & {\bf 25.4} & {\bf 12.7} & {\bf 6.4}  &  {\bf 13.0}  & {\bf 31.3} & {\bf 29.0} & {\bf 9.1} \\ \hline

      \multirow{2}*{Flickr30k$\leftrightarrow$MSCOCO} &SME-GAN~\cite{laina2019towards}  &     -      &     -      &     -      &    7.9     &   13.0   &    32.8    &    9.9     &    7.5    \\
                    ~ &$R^2M$                 & {\bf 53.1} & {\bf 32.8} & {\bf 19.2} & {\bf 11.7} &  {\bf 13.7 } &    {\bf 35.9}    & {\bf 18.1} &   {\bf  8.3 }   \\ \hline

      \multirow{2}*{MSCOCO$\leftrightarrow$GCC} &SME-GAN~\cite{laina2019towards}  &     -      &     -      &     -      &    6.5     &  12.9  & {\bf 35.1} &    22.7    &    7.4    \\
                    ~ &$R^2M$                  &   {\bf  51.2}    &   {\bf  29.5 }   &   {\bf  15.4 }   &   {\bf  8.3 }    & {\bf  14.0}  &    35.0   & {\bf 29.3} & {\bf 9.6} \\ \bottomrule[1pt]
   \end{tabular}}
   \caption{Performance comparison with the state-of-the-art methods. The best performance is marked with bold face.}
\label{table_2}
\end{table*}

\section{Experiments}
\label{expe}

\subsection{Dataset and Metrics}
\label{exp_data}
We test all the existing unsupervised image captioning datasets, including (1) MSCOCO images~\cite{lin2014microsoft} paired with Shutterstock captions \cite{feng2019unsupervised}; and (2) Flickr30k images \cite{young2014image} paired with MSCOCO captions and (3) MSCOCO images paired with Google’s Conceptual Captions (GCC) \cite{sharma2018conceptual,laina2019towards}.
In the test splits of datasets, each image has five
ground-truth captions. 

\subsection{Implementation Details}
We split each image set and filter captions as \cite{feng2019unsupervised,laina2019towards}. The visual dictionary $\mathcal D$ in Fig.\ref{fig_3} is collected by a pre-trained Faster R-CNN \cite{huang2017speed}
OpenImages-v4 \cite{krasin2017openimages,kuznetsova2018open}. We merge the visual concepts in $\mathcal D$ and words in training captions into a large vocabulary, to cover the majority of the to-be-generated words. The vocabulary sizes of the three datasets are 18,679/11,335/10,652, respectively, including tokens $<$\#start$>$, $<$\#end$>$, and $<$UNK$>$.
For experimental setting, we filter out visual concepts form images with the detected score $\geq0.3$. Both the sizes of LSTM and RM memory are set to $N=1$ and $d=512$. The parameters of multi-head self-attention are $H=2$, $d_k=d_{\mathcal{K}}=256$, and $d_v=d_{\mathcal{V}}=256$. The margin in Eq.~\ref{eq:12} is $\mathbf m=0.2$. Adam optimizer is adopted 
with batch size of 256. For three datasets, hyper-parameters ($\beta$, $\gamma$) are set to (1, 1), (1, 1), (0.2, 0.2). We train the model with a loss $L\!_{X\!E}$ under learning rate 10$^{-4}$, while fine-tune it with the joint loss $L_{S}$. After that, $L^I_M$ is used to train with a learning rate 10$^{-5}$. Finally, we jointly train the model with $L_{I}$.
In the test process, we use the beam search tactic~\cite{anderson2016guided} with width of 3.

\begin{table}
   \setlength{\tabcolsep}{.3em}
   \centering
   \begin{threeparttable}
   \scalebox{0.66}{
   \begin{tabular}{c|c|cccccccc}
      \toprule[1pt]
            Dataset &Method  &  B-1   &  B-2   & B-3  &   B-4   & M &  R   & C   &   S  \\ \midrule
      \multirow{4}*{MSCOCO$\leftrightarrow$} & D w/o FM &  33.0        &  19.0       &   9.6      &   4.9     &  10.5   &   26.9      &   23.7     &   7.6     \\
      ~   &D w/o Memory in RM& 39.9         &   22.3      &   10.8      &   5.2     &  12.1   &   28.9      & 25.9       &  8.4      \\
      \multirow{2}*{Shutterstock}   &R w/o Memory in RM &  38.8        &  21.6       &  10.6       &   5.1     &  10.8   &  26.8       &  24.5      &  8.0       \\
      ~   &D\&R w/o Memory in RM &  40.5     & 22.3        & 10.6      &   5.2     &  12.2  &   28.8      &  25.9      &  8.5    \\
      ~   &$R^2M$                 &    {\bf 44.0 }   & {\bf 25.4} & {\bf 12.7} & {\bf 6.4}  &  {\bf 13.0}  & {\bf 31.3} & {\bf 29.0} & {\bf 9.1} \\ \hline
      \multirow{4}*{Flickr30k$\leftrightarrow$} &D w/o FM &  52.9        &  32.5       &  18.8       &  11.2      & 13.0    &   35.5      &   14.9     &   8.0     \\
      ~   &D w/o Memory in RM &52.5          & 32.4        &  19.0       &  11.6      &  13.0   &  35.5       &  16.1      &  7.8      \\
      \multirow{2}*{MSCOCO}   &R w/o Memory in RM & 51.5         &  30.9       &  17.5       &  10.3      &  13.3   &  35.0       &  16.3      &  7.9      \\
      ~   &D\&R w/o Memory in RM & 52.2      &   31.8      &  18.5     &  11.1      & 13.2   &  35.6       &  16.0      &  8.1    \\
      ~   &$R^2M$                 & {\bf 53.1} & {\bf 32.8} & {\bf 19.2} & {\bf 11.7} &  {\bf 13.7 } &    {\bf 35.9}    & {\bf 18.1} &   {\bf  8.3 }   \\ \hline
      \multirow{4}*{MSCOCO$\leftrightarrow$} &D w/o FM &   39.1       &  23.4       &   12.2      &  6.7      &  11.3   &   32.4      &   23.9     &    7.8     \\
      ~   &D w/o Memory in RM &  43.0        &  24.8       &   12.5      &  6.7      &  12.4   &   32.5      &  27.1      &  8.8      \\
      \multirow{2}*{GCC}   &R w/o Memory in RM &  47.2        &  26.9       &   13.9      &  7.3      & 12.9    &  33.2       &  28.1      &  9.0      \\
      ~   &D\&R w/o Memory in RM &  43.6     &   25.8      &   13.4    &   7.3     & 12.5   &  33.3       &  27.8      &  8.7    \\
      ~   &$R^2M$                &   {\bf  51.2}    &   {\bf  29.5 }   &   {\bf  15.4 }   &   {\bf  8.3 }    & {\bf  14.0}  &    {\bf 35.0}   & {\bf 29.3} & {\bf 9.6} \\ \bottomrule[1pt]

   \end{tabular}}
   \end{threeparttable}

      \caption{Ablation studies of $R^2M$ with different memory settings. The best performance is marked with bold face. \textcircled{1} In ``$D$ w/o FM", ${\bm f}_{t}$ is calculated by a linear layer on the concatenation of ${\bm v}$ and ${\bm w}$. \textcircled{2} ``$D\&R$ w/o Memory in RM" replaces the RM operation by LSTM in both $R^2M. Decoder$ and $Reconstructor$. \textcircled{3} ``$D$ w/o Memory in RM" and \textcircled{4} ``$R$ w/o Memory in RM" replaces RM by LSTM in respective $R^2M. Decoder$ and $Reconstructor$.}
\label{table_4}
\end{table}

\subsection{Experimental Results and Analysis}
\subsubsection{Comparison with the State-of-the-arts} $R^2M$ exhibits large improvements across all the metrics. Both UC-GAN~\cite{feng2019unsupervised} and SME-GAN~\cite{laina2019towards} rely on complicated GAN training strategies, whereas ours $R^2M$ is a memory solution. As shown in Table \ref{table_2}, $R^2M$ upgrades BLEU-4 (B-4) by 14.3\%, 48.1\% and 27.7\% on three datasets, where BLEU-4 involves 4-gram phrases. It implies the stronger capacity of $R^2M$ to learn long-range dependencies than others. 
$R^2M$ also raises CIDEr/SPICE, from 28.6/8.1 to (29.0/9.1), 9.9/7.5 (18.1/8.3) and 22.7/7.4 (29.3/9.6). The promising improvements demonstrate the consistency of superior performances. 
With the released code of UC-GAN~\cite{feng2019unsupervised} on the
MSCOCO$\leftrightarrow$Shutterstock dataset, here is an efficiency comparison: $R^2M$ vs. UC-GAN $\approx$ 35 $min$ vs. 34 $hours$. 
$R^2M$ also enjoys higher computational efficiency.

\subsubsection{Ablation Study of $R^2M$}
To verify each component in $R^2M$, we propose the ablation study.
\textbf{(1) Effect of FM.} Compared to the entire $R^2M$, the performance of ``$D$ w/o FM'' drops significantly, \emph{e.g.}, with 18.3\%, 17.7\% and 18.4\% reduction of CIDEr (C) on three datasets in Table \ref{table_4}. FM effectively implements the implicit correlation between visual concept vector and word embedding. 
\textbf{(2) Effect of memory in RM.} For Table \ref{table_4}, either ``$D$, $R$ or $D\&R$ w/o Memory in RM'' suffers from worse performance, \emph{e.g.}, on dataset MSCOCO$\leftrightarrow$GCC, dropping the CIDEr from 29.3 to 27.1, 28.1 and 27.8. RM excels at storing and retrieving information across time than classical LSTM, to effectively handle sequential learning.
\textbf{(3) Effect of Loss.} In each block diagram of Table \ref{table_3}, the first line records the result of model trained with only $L\!_{X\!E}$ on text corpus. Note that this baseline is competitive and outperforms the existing methods. The SPICE (S) is increased by around 11.1\%, 9.3\% and 25.7\% compared on three datasets.
By gradually incorporating $L_{rec}^{S}$, $L\!_{M}^{I}$ and $L_{rec}^{I}$, the model performs much better. The CIDEr gradually raises from 25.4 to 27.0, 28.9 and 29.0 on MSCOCO$\leftrightarrow$Shutterstock. Especially after the assistance of semantic matching loss $L\!_M$, the CIDEr is significantly improved, nearly 7.0\%, 6.6\% and 3.2\% on all datasets. 

\begin{table}[t]
   \setlength{\tabcolsep}{.3em}
   \centering
   \scalebox{0.66}{
   \begin{tabular}{c|p{0.6cm}p{0.5cm}p{0.4cm}p{0.6cm}|cccccccc}
      \toprule[1pt]
            Dataset &$L\!_{X\!E}$ & $L_{rec}^{S}$ &$L\!_{M}^{I}$ & $L_{rec}^{I}$  &  B-1  &  B-2   &  B-3   & B-4  & M &   R    &   C   &  S \\  \midrule

                  \multirow{3}*{MSCOCO$\leftrightarrow$}  & $\checkmark$ &               &         &         &    42.2     &    23.2    &    11.3     &     5.7     &     13.0     &    29.2     &    25.4    &    9.0    \\
               ~   & $\checkmark$ & $\checkmark$  &        &          &    {\bf  44.7}     &    25.0    &    12.2     &     6.0     &     {\bf  13.3}     &    30.7     &    27.0    &    9.1    \\
                    \multirow{1}*{Shutterstock}      & $\checkmark$ & $\checkmark$  &$\checkmark$        &          &    44.2     &    25.4    &    12.7     &     6.3     &     13.1     &    31.3     &    28.9    &    9.1   \\
                     ~   & $\checkmark$ & $\checkmark$  & $\checkmark$  & $\checkmark$ & 44.0 & {\bf 25.4} & {\bf 12.7}  &  {\bf 6.4}  & 13.0  & {\bf 31.3}  & {\bf 29.0} & {\bf 9.1}  \\ \hline

                  \multirow{3}*{Flickr30k$\leftrightarrow$} & $\checkmark$ &               &       &           &    49.9     &    30.0    &    17.1     &     10.1     &     13.5     &    34.9     &    16.4    &    8.2    \\
                ~ & $\checkmark$ & $\checkmark$  &        &          &    49.4     &    29.8    &    17.4     &     10.5     &     13.8     &    35.1     &    16.7    &    8.2    \\
                      \multirow{1}*{MSCOCO}  & $\checkmark$ & $\checkmark$  &$\checkmark$        &          &    51.8     &    31.6    &    18.4     &     11.0     &       {\bf 13.9 }     &    35.7     &    17.8    &    8.3   \\
                      ~ & $\checkmark$ & $\checkmark$  & $\checkmark$ & $\checkmark$   & {\bf 53.1} & {\bf 32.8} & {\bf 19.2} & {\bf 11.7} &  13.7 &    {\bf 35.9}    & {\bf 18.1} &   {\bf  8.3 }  \\ \hline

                  \multirow{3}*{MSCOCO$\leftrightarrow$}  & $\checkmark$ &               &         &         &    46.4     &    25.8    &    12.8     &     6.7     &     13.9     &    32.6     &    26.9    &    9.3    \\
                ~  & $\checkmark$ & $\checkmark$  &         &         &    49.2     &    27.9    &    14.3     &     7.7     &     13.6     &    33.6     &    28.2    &    9.3    \\
                      \multirow{1}*{GCC} & $\checkmark$ & $\checkmark$  & $\checkmark$      &           &    51.0     &    29.3    &    15.3     &    {\bf 8.4}    &     13.9     &    34.8     &    29.1    &    9.6   \\
                      ~ & $\checkmark$ & $\checkmark$  & $\checkmark$ & $\checkmark$   & {\bf  51.2} & {\bf 29.5} & {\bf 15.4}  &  8.3 & {\bf  14.0}  & {\bf 35.0}  & {\bf 29.3} & {\bf 9.6}  \\ \bottomrule[1pt]
   \end{tabular}
   }
   \caption{Ablation studies of $R^2M$ with different losses. The best performance is marked with bold face.}
\label{table_3}
\end{table}

\begin{figure*}[t]
  \centering
  \includegraphics[width=\linewidth]{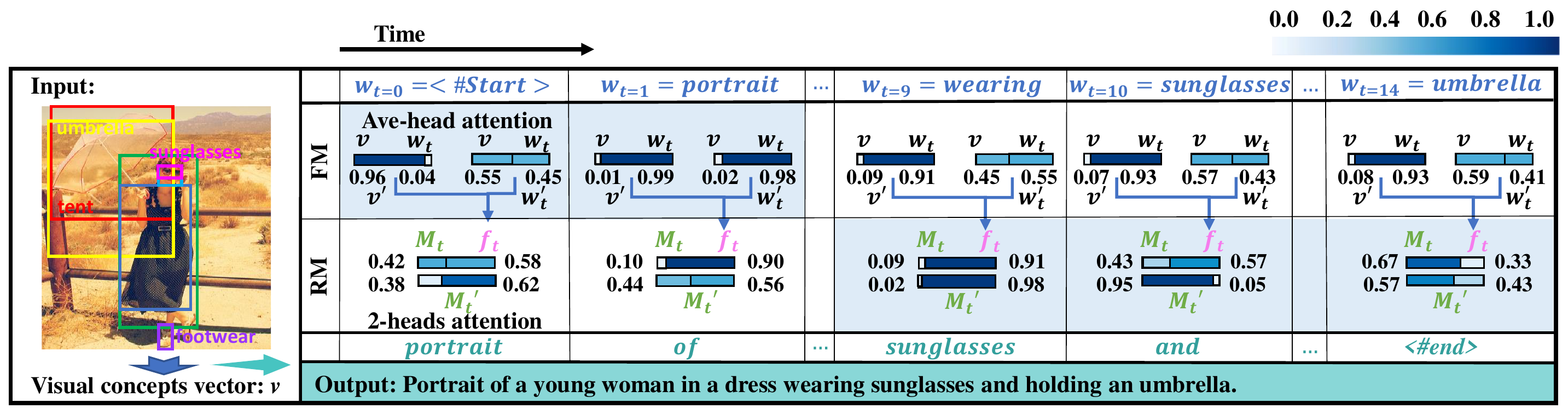}
  \caption{Visualization results of attention weights in FM and RM units of $R^2M. Decoder$. The weights reflect how much attention the model pays to each input variable.}
\label{fig_6}
\end{figure*}

\begin{figure}[h]
   \centering
   \includegraphics[width=\linewidth]{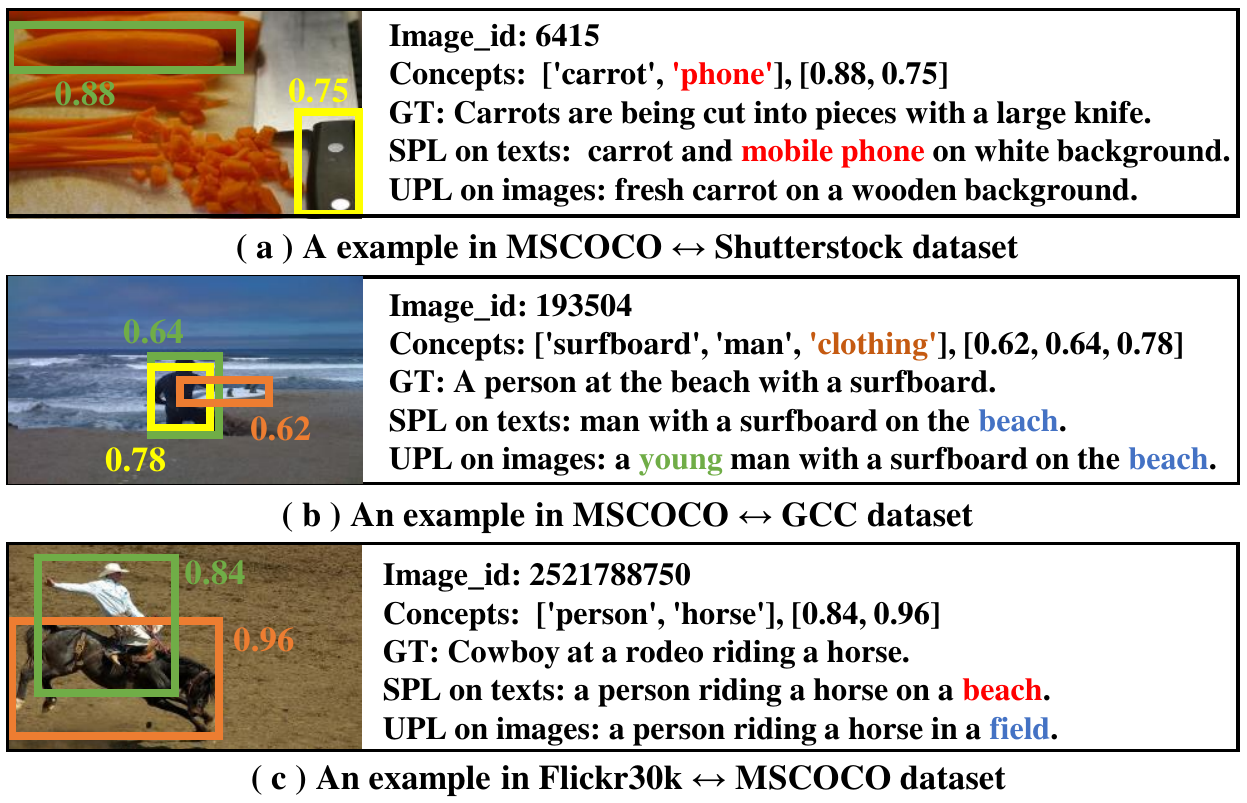}
   \caption{Qualitative examples of $R^2M$. Visual concepts are detected by Faster R-CNN. Words in $\{ red, brown, blue, green \}$ fonts mark the incorrectly recognized, irrelevant, and new detected concepts, and the adjective, respectively. 
   }
\label{fig_7}
\end{figure}

\begin{figure}[t]
   \centering
   \includegraphics[width=\linewidth]{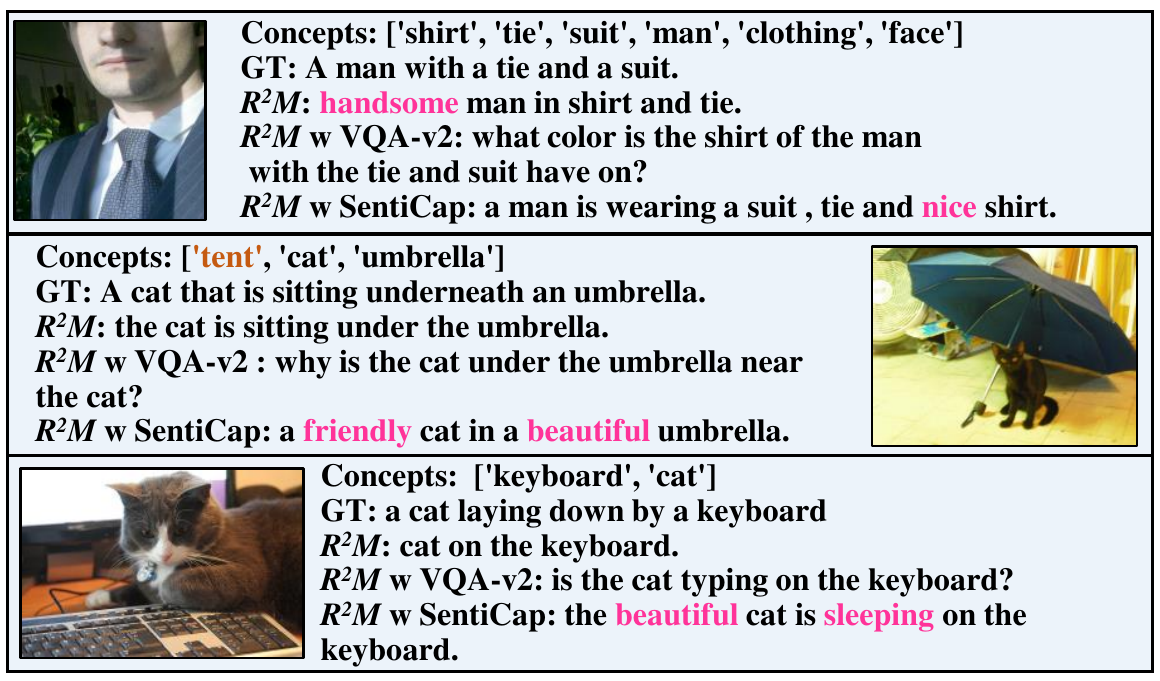}
   \caption{Extensive captions of MSCOCO images in different language styles. Words in $\{ brown, pink \}$ fonts highlight the incorrect detected concepts and the sentiment adjectives, respectively.}
\label{fig_8}
\end{figure}

\subsubsection{Qualitative Results}
\textbf{Visualization of Attention Weights in FM \& RM.} Fig.\ref{fig_6} illustrates an example of memory learning in $R^2M. Decoder$, which is interpretable. FM displays the average weight of multi-head attention, while RM offers $H=2$ heads attention. With the beginning token $<$\#start$>_{\{t=0\}}$ and the encoded concept vector ${\bm v}$, FM pays more attention to the richer semantics ${\bm v}$. And at time $t=2$, FM focuses much more on the previous word $portrait_{\{t=1\}}$ as $portrait$ is the first generated concept and deserves more attention.
Then, we discuss the interpretation of RM. Taking previous word $wearing_{\{t=9\}}$ as an example, it affects the generation of $sunglasses_{\{t=10\}}$ more influentially than the memory $M^d_t$. However, at $t=11$, under the previous cue $sunglasses_{\{t=10\}}$, the model infers a relational conjunction $and_{\{t=11\}}$ by mainly recalling $M^d_t$. The same situation holds at the last time $umbrella_{\{t=14\}}$, there is no relevance cues to be found from $M^d_t$, the model decides to terminate the entire generation process.
\newline
\textbf{Visualization of Generated Captions.} We detect visual concepts and their scores by Faster R-CNN. As shown in Fig.\ref{fig_7} (a), $phone$ is an incorrectly detected object with a high score 0.75. While performing training on text corpus with $L\!_{X\!E}$ and $L^S_{rec}$, $R^2M$ translates discrete concepts to a sentence, still containing $phone$.
With further unsupervised training on images over $L^I_M$ and $L^I_{rec}$, $R^2M$ automatically eliminates the wrong concept. By contrast, exemplified in Fig.\ref{fig_7} (b), $clothing$ is a correctly identified concept, but irrelevant to salient visual regions of the image. $R^2M$ eliminates the redundant visual concepts yet. Moreover, there are new learned concepts $beach$ and an adjective $young$ from all the joint SPL and UPL semantic learning. To strengthen the intuition that $R^2M$ can extrapolate beyond the concepts in the images, we offer another example in Fig.\ref{fig_7} (c). Both the new words $beach$ and $field$ are undetected visual concepts. Following the textual cues learning from text corpus, $R^2M$ acquires the knowledge to infer a new context-independent concept $beach$; however, it is irrelevant. After unsupervised visual alignment learning, the caption finally outputs a new word $field$ instead of $beach$. $R^2M$ is effective to infer promising descriptions about images without annotated captions.

We also extend the experiments with new corpora with \textbf{different language styles}, such as VQA-v2 \cite{antol2015vqa} and SentiCap \cite{mathews2016senticap}, involving the questions about the visual content and sentiment captions. For our experiments, 1,105,904 questions provided by VQA-v2 and 4,892 positive captions of SentiCap are respectively trained as text corpus. As shown in Fig.\ref{fig_8}, $R^2M$ also excels at questioning and describing images with positive emotion.

\section{Conclusion}
\label{conc}
This paper proposes a novel recurrent relational memory network ($R^2M$) for unsupervised image captioning with low cost of supervision. $R^2M$ is a lightweight network, characterizing self-attention and a relational gate to design the fusion and recurrent memory for long-term semantic generation. Experimental results show that the $R^2M$ surpasses the state-of-the-arts on three
benchmark datasets.

\section*{Acknowledgments}
This work is supported by the National Key Research and Development Program of China under grant 2018YFB0804205, and the National Natural Science Foundation of China (NSFC) under grants 61806035, U1936217, 61725203, 61732008, and 61876058.
\bibliographystyle{named}
\bibliography{ijcai20}

\end{document}